\pdfoutput=1
\documentclass[letterpaper]{article}
\usepackage{aaai20}
\usepackage{times}
\usepackage{helvet}
\usepackage{courier}

\usepackage{amsfonts}
\usepackage{siunitx}
\usepackage{multirow}
\usepackage{enumerate}
\usepackage{booktabs}
\usepackage{amssymb}
\usepackage{calc}
\usepackage{paralist}
\usepackage[labelfont=bf,textfont=bf,singlelinecheck=on]{subcaption}
\usepackage{tikz}
\usetikzlibrary{fit,positioning}
\usepackage{graphicx}
\usepackage[justification=justified]{caption}
\usepackage{diagbox}
\usepackage{color}
\usepackage{colortbl}
\usepackage{url}
\usepackage{mathtools}
\usepackage{algorithm}
\usepackage[normalem]{ulem}
\usepackage{enumitem}
\usepackage{numprint}
\npthousandsep{,}
\npdecimalsign{.}
\usepackage[skip=0pt]{caption}
\usepackage{setspace}
\usepackage{color}
\usepackage{colortbl}
\usepackage{acronym}

\acrodef{ED}{event detection}
\acrodef{FSED}{few-shot event detection}
\acrodef{EC}{event classification}
\acrodef{MLM}{masked language model}
\acrodef{NLP}{natural language processing}
\acrodef{PLMs}{pretrained language models}
\acrodef{MsPrompt}{multi-step prompt}
\acrodef{CNN}{convolutional neural network}
\acrodef{RNN}{recurrent neural network}
\acrodef{GNN}{graph neural network}

\newcommand{\citet}[1]{\citeauthor{#1} \shortcite{#1}}
\newcommand{\citep}{\cite}

\title{MsPrompt: Multi-step Prompt Learning for Debiasing Few-shot Event Detection}
\author{Siyuan Wang$^1$\qquad Jianming Zheng$^1$\qquad Xuejun Hu$^2$\qquad Fei Cai$^1$\qquad Chengyu Song$^1$\qquad Xueshan Luo$^1$\\
	$^1$~Science and Technology on Information Systems Engineering Laboratory\\ National University of Defense Technology, Changsha, China\\
	$^2$~Business School, Hunan University, Changsha, Hunan, China
	\\{wangsiyuan21, zhengjianming12, caifei08, songchengyu, luoxueshan}@nudt.edu.cn, xuejun\_hu@hnu.edu.cn}

\begin{document}
\maketitle

\begin{abstract}
	Event detection (ED) is aimed to identify the key trigger words in unstructured text and predict the event types accordingly.
	Traditional ED models are too data-hungry to accommodate real applications with scarce labeled data.
	Besides, typical ED models are facing the context-bypassing and disabled generalization issues caused by the trigger bias stemming from ED datasets.
	Therefore, we focus on the true few-shot paradigm to satisfy the low-resource scenarios.
	In particular, we propose a \textbf{m}ulti-\textbf{s}tep \textbf{prompt} learning model (MsPrompt) for debiasing few-shot event detection, that consists of the following three components:
	an under-sampling module targeting to construct a novel training set that accommodates the true few-shot setting, 
	a multi-step prompt module equipped with a knowledge-enhanced ontology to leverage the event semantics and latent prior knowledge in the PLMs sufficiently for tackling the context-bypassing problem, 
	and a prototypical module compensating for the weakness of classifying events with sparse data and boost the generalization performance.
	
	Experiments on two public datasets \emph{ACE-2005} and \emph{FewEvent} show that MsPrompt can outperform the state-of-the-art models, especially in the strict low-resource scenarios reporting 11.43\% improvement in terms of weighted F1-score against the best-performing baseline and achieving an outstanding debiasing performance.
\end{abstract}

\section{Introduction}
\label{Introduction}

As a cornerstone of natural language processing, \acf{ED} supports numerous downstream tasks, e.g., event extraction~\citep{kbs22/Liu,acl22/Wang,ipm21/Zhang}, text classification~\citep{access/Liu22d,Zheng2020Pre}, information retrieval~\citep{www22/Madisetty,ictir21/Voskarides,icmlc20/Zhao,cikm20/Zhao},dialogue recognition~\citep{coling22/Wei}, etc.
However, due to the data-hungry trait of deep learning, traditional ED models often struggle in the scenario where annotating sufficient labeled data is unaffordable.
In this light, \acf{FSED} \citep{Zheng2021Tax} is proposed, which aims at making predictions with few labeled data.

Recently, FSED work achieves much progress by the virtue of meta learning \citep{iccv03/Fei}, which trains and captures the meta knowledge of event type on the tremendous labeled data of old event types, thereby helping FSED models generalize quickly to novel event types with scarce labeled data. 
These FSED approaches can mainly be classified into metric-based FSED methods \citep{emnlp21/Lai,acl21/Cong} and optimization-based FSED ones \citep{sigir21/Lai}.
However, such ambitious data demands in old event types cannot be satisfied completely in some application scenarios.
In this paper, we therefore focus on the true few-shot training setting (i.e., only providing a few labeled data regardless of old or new event types) \citep{NeurIPS21/Perez} for event detection.

\begin{figure}[t]
\centering
\includegraphics[width=1.0\columnwidth]{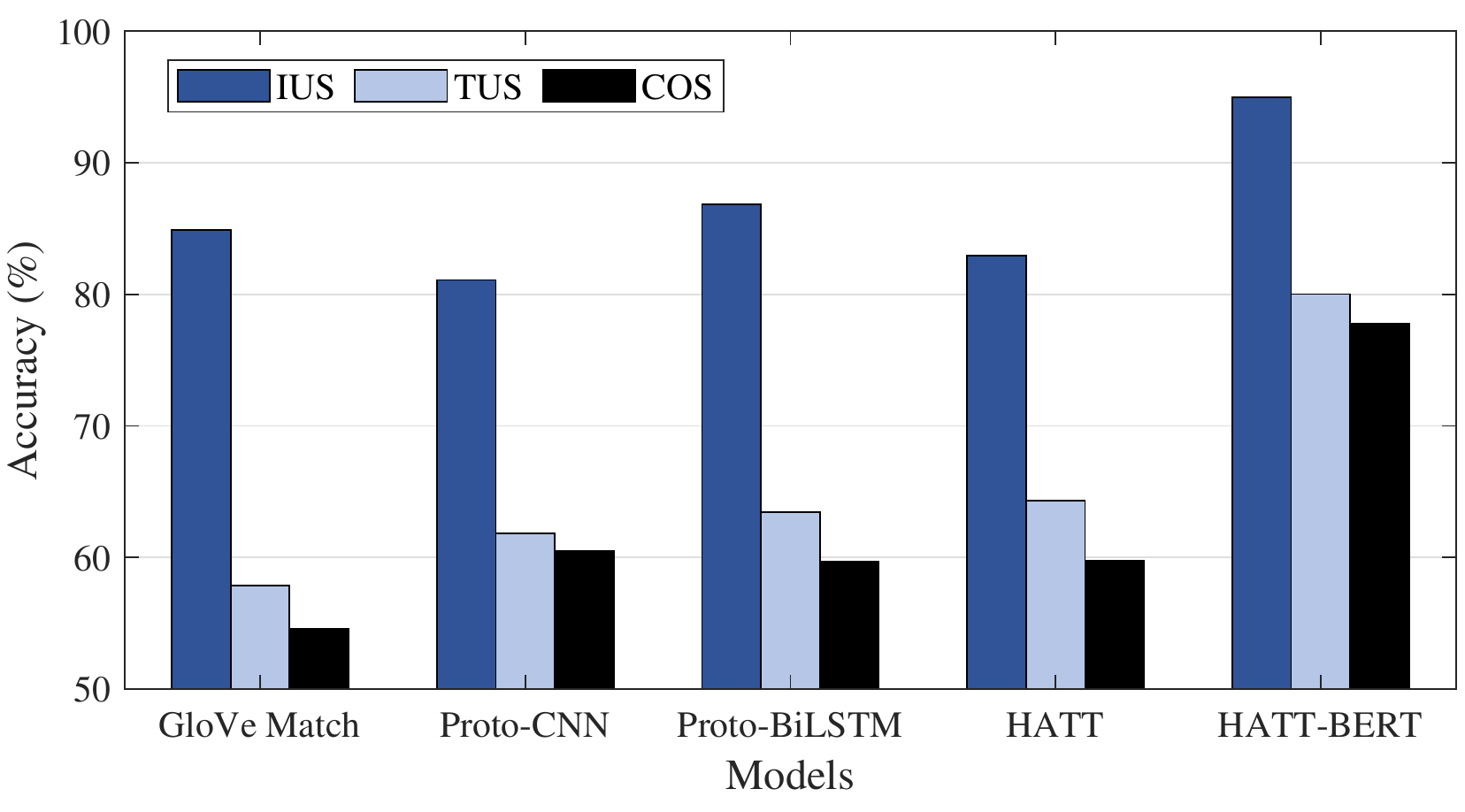}
\caption{Model performance on \emph{FewEvent} under different sampling methods, where the \textbf{I}nstance \textbf{U}niform \textbf{S}ampling (IUS) produces the original trigger-biased test set, while the \textbf{T}rigger \textbf{U}niform \textbf{S}ampling (TUS) and \textbf{CO}nfusion \textbf{S}ampling (COS) construct the test sets without the trigger bias, respectively. These results are produced by \citet{cikm21/Wang}.}
\label{Figure0}
\end{figure}

In addition, existing FSED approaches are typically exposed to the trigger bias stemming from the ED datasets.
Taking a look at the \emph{FewEvent} dataset \citep{wsdm20/Deng}, whether the trigger words of any event type or the event types triggered by the same word, their frequencies strictly follow the long-tail distribution.
For instance, for the event type ``\emph{Life.Marry}", the percentage of its top-3 trigger words reaches 64.97\%. 
While for the trigger word ``\emph{work}", the percentage of top-3 event types triggered by this word amounts to 99.53\%.
Such long-tail distributions will easily lead to the following issues for event detection. 
The first is the context-bypassing issue. 
In the few-shot scenario, the FSED models are easily tempted to have over-confidence in high-frequency trigger words or event types, thereby simply taking the trigger words as clues to determine the event types without considering any event context.
The second is the generalization disability issue.
Due to the scanty evidence brought by the context-bypassing issue, the FSED models cannot generalize to low-frequency trigger words or event types.
These aforementioned issues can be further validated by the performance comparison on biased and unbiased test sets. 
As shown in Fig.\ref{Figure0}, the performance on unbiased test sets (i.e., TUS and COS) is drastically lower than that on the biased one (i.e., IUS).
Such performance drops indicate that the trigger bias in the ED datasets can make the FSED models obtain inflated performance.
Similar phenomena can also be found on the \emph{ACE-2005} dataset \citep{lrec04/Doddington}.

In this paper, we attempt to accommodate the low-resource scenarios and overcome the trigger bias in few-shot event detection by proposing a \textbf{m}ulti-\textbf{s}tep \textbf{prompt} learning model (MsPrompt), which consists of three main components, including an under-sampling module, a multi-step prompt module, and a prototypical module.
In particular, the under-sampling module aims to construct the training data abiding by the true few-shot format.
Under such a radical data setting, recent popular prompt learning ~\citep{NeurIPS20/brown,ACL21/Gao} which elicits the latent knowledge from the \acf{PLMs} by some prompts can provide sufficient latent information for the FSED models and thus forms the multi-step prompt module.
This module extends the existing one-step prompt to a multi-step one, which refines the FSED process into two consecutive subprompts, i.e., the trigger recognizer and the event classifier.
More specifically, such two subprompts resort to the PLMs to locate the trigger word and predict the event type, where the generation format forces the FSED models to concentrate on the event context, thereby mitigating the context-bypassing risk.
In addition, this module introduces the knowledge-enhanced ontology to enrich the prompt semantics.
Finally, the prototypical module, dealing with the generalization disability issue, employs the prototypical networks \citep{nips17/Snell} to obtain the representation of event types by clustering, which removes the noise of high-frequency labels and enhances the discrimination ability of low-frequency labels. 

Extensive experiments for few-shot event detection are conducted on 
\emph{ACE-2005} and \emph{FewEvent}. 
The results show that our proposed MsPrompt achieves obvious improvements in terms of weighted F1-score over the state-of-the-art baselines in the strict low-resource scenarios.
Moreover, the evaluations on the biased and unbiased test sets verify the debiasing potentiality and strong generalization of MsPrompt.

Our key contributions in this paper can be summarized in the following three folds.
\begin{enumerate}[leftmargin=*,nosep]
\item To address the context-bypassing problem, we extend the one-step prompt to a multi-step one by disassembling the FSED process into two consecutive subprompts, which can efficiently aggregate the predicted trigger, the knowledge-enhanced ontology, and the latent knowledge in PLMs to focus on the context and predict accurately.

\item We design a novel prototypical module to mitigate the disabled generalization issue by strengthening the discrimination of low-frequency labels.

\item We verify the effectiveness of MsPrompt against the state-of-the-art baselines on \emph{ACE-2005} and \emph{FewEvent} for the true few-shot event detection and find that MsPrompt can not only achieve better model performance but make progress in debiasing triggers.
\end{enumerate}

\section{Related Work}
\label{Related Work}
We review the related works from two main aspects, i.e., event detection and debiasing methods.

\subsection{Event detection}
\label{Event detection}
According to the accessibility of data resources, the task of event detection can be roughly divided into two categories: data-rich and few-shot ED.
In data-rich ED, sufficient data provides a guarantee for the traditional neural networks, such as \acf{CNN} \citep{ACL15/chen,ACL15/nguyen}, \acf{RNN} \citep{NAACL16/jagannatha,NAACL16/nguyen}, \acf{GNN} \citep{emnlp18/Liu,aaai18/Nguyen,TPAMI22/peng}, and self-attention network \citep{aaai18/Liu,acl17/Liu}.
In particular, to reduce the cost of tagging triggers, \citet{naacl19/Liu} propose a type-aware bias neural network to encode a sentence with the target event types. 
\citet{aaai22/Xie} develop an approach based on Graph Convolutional Network (GCN) for information extraction and aggregation on the graph to alleviate the heavy reliance on a fixed syntactic parse tree structure.

Few-shot ED aims to alleviate the problems such as generalization bottlenecks caused by insufficient data and maintain the excellent detection performance in low-resource scenarios.
\citet{wsdm20/Deng} construct a dynamic-memory-based prototypical network to generate a robust event representation through the multi-hop mechanism.
\citet{acl21/Cong} propose the prototypical amortized conditional random field to handle the label dependency issue. 
\citet{sigir21/Lai} introduce a regulating representation based on graphs and word sense disambiguation to improve the generalization performance.
\citet{Zheng2021Tax} model the taxonomy-aware distance relations and integrate the Poincaré embeddings into a TapNet.

However, either data-rich or few-shot ED requires abundant held-out classes to achieve high performance, which is divorced from most application scenarios in which labeled data is difficult to obtain.
Therefore we follow the true few-shot training format, i.e., a small validation set of the same size as the few-shot training set.
To capture event information with extremely sparse data, we further employ a prompt-based method for the true few-shot event detection, which can drive huge PLMs and evoke the inherent knowledge therein to enrich the semantics of event representation.

\subsection{Debiasing methods}
\label{Debiasing methods}
As predictive biases emerge in various tasks of NLP, two serious consequences, i.e., outcome disparities and error disparities, can not be ignored ~\citep{acl20/Shah}.
Therefore, we summarize a series of debiasing methods in detail below.

\subsubsection{Debiasing methods in NLP}
\label{Debiasing methods in NLP}
According to the bias source, \citet{acl20/Shah} divide the biases in NLP into four categories: label bias, selection bias, model over-amplification, and semantic bias.
To release the label bias attributed to the incorrect annotation, \citet{acl21/Chen} design a counterfactual-based debiasing framework for text classification.
For the selection bias, i.e., the phenomenon in which the training observations differ from the target distribution due to the non-representative training data, \citet{acl21/Liu} propose an additional saliency selection layer and an optimization method to mitigate the implicit bias of deep text classifiers.
\citet{icml21/Zhao} observe the majority label bias and recency bias existing in the prompt model of GPT-3, both of which are the over-amplifying bias.
That is, models rely on imperfect evidence for predictive shortcuts.
~\citet{acl22/Guo} automatically mitigate the semantic biases, embodied in undesired social stereotypes carried by PLMs.

\subsubsection{Debiasing methods in ED}
\label{Debiasing methods in ED}
As mentioned in Section~\ref{Introduction}, the severe trigger bias is widely appeared in event detection, which essentially belongs to the selection bias and over-amplifying bias \citep{song2022augprompt}.
To address the context-bypassing problem caused by the trigger bias, \citet{cikm21/Wang} employ adversarial training and trigger reconstruction techniques to ease the over-reliance on the trigger.
Although the context-bypassing is alleviated, in few-shot occasions, the addition of noise to the trigger embedding may induce the misclassification of the model, thus aggravating the disabled generalization.
\citet{acl20/Tong} provide an enrichment knowledge distillation model to reduce the inherent bias of common trigger words by inducting open-domain trigger knowledge.
However, a large number of unlimited candidate triggers from unlabeled data imported by open-domain knowledge cause a great interference to the event detection model and harm the generalization performance, especially for FSED.
In addition, \citet{emnlp21/Chen} perform a causal intervention on the context by a backdoor adjustment to mitigate overfitting induced by the trigger bias.

Albeit much progress, these debiasing strategies often ignore the context-bypassing or disabled generalization issue, and thus cannot be applied to real low-resource scenarios.
We argue these two issues should be dealt with jointly to improve the prediction and generalization performance.
Hence, we introduce a task-oriented and knowledge-enhanced ontology without adding noise and develop a novel debiasing proposal MsPrompt for few-shot event detection.

\section{Approach}
\label{Approach}
In this section, we first formulate the task of event detection and detail the MsPrompt model in Section~\ref{Task formulation and model framework}, which consists of three main components, including an under-sampling module (see Section~\ref{Under-sampling}), a multi-step prompt module (see Section~\ref{Multi-step Prompt}), and a prototypical module (see Section~\ref{Prototypical network}).

\begin{figure*}[h]
\centering
\includegraphics[width=1.0\textwidth]{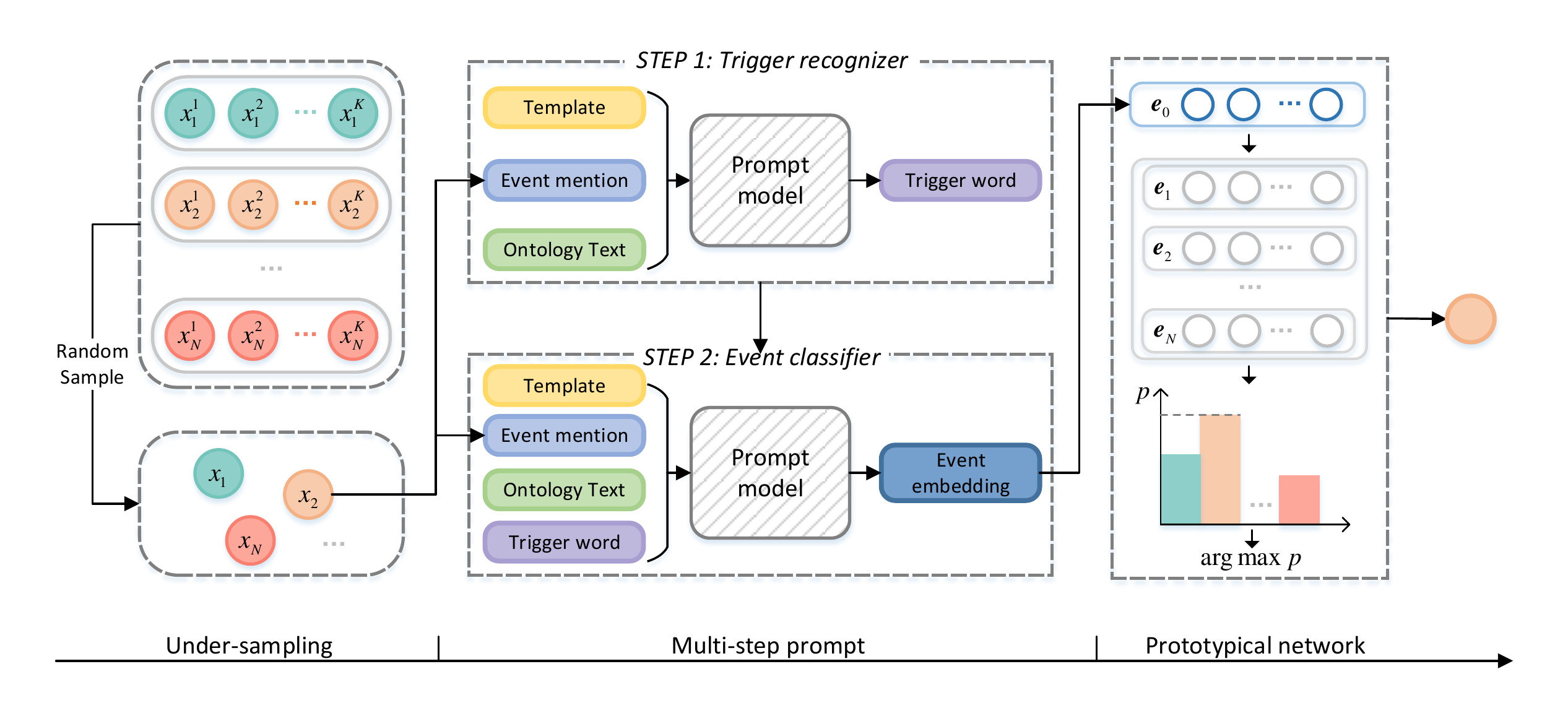}
\caption{The workflow of MsPrompt.}
\label{Figure2}
\end{figure*}

\subsection{Task formulation and model framework}
\label{Task formulation and model framework}
\noindent\textbf{ED}.
The task of event detection can be formulated as $x \to (x, t) \to y$, where $x$ represents the event mention, and $y \in Y$ represents the event type predicted from the predefined event label set $Y$.
In the intermediate step, the trigger $t$ is identified from the input $x$, in the form of a word or a phrase which triggers the corresponding event label $y$ \citep{rep4nlp16/Nguyen}.

\noindent\textbf{Few-shot ED}.
Following the archetypal $N$-way $K$-shot training format, a meta task is constructed by a support set $\mathcal{S}$ with $N$ novel event labels that contain $K$ instances per label and a query set $\mathcal{Q}$ that includes unlabeled instances to be predicted from $\mathcal{S}$.
Typically, a meta task can accurately detect the event type of each query instance from $\mathcal{S}$ with only a few labeled data.
However, few-shot ED requires a large amount of old event label data to extract the meta knowledge, making it unsuitable for the real scenarios.

\noindent\textbf{True few-shot ED}.
Formally, given a group of instances with their corresponding trigger $t$ and event label $y \in Y$, each label contains $K$ instances, making up the true few-shot training set.
True few-shot ED, typically trained from the true few-shot training set and evaluated by a validation set of the same size, targets to identify the trigger word and detect the predefined event label in the low-resource scenarios.

\noindent\textbf{Framework of MsPrompt}.
Based on the true few-shot ED formulation, we propose a novel approach MsPrompt.
The workflow of MsPrompt is shown in Fig.~\ref{Figure2}.
First, through an under-sampling module, a data matrix $\textbf{X} \in \mathbb{R}^{N \times K}$ is constructed from the initial annotated dataset to satisfy the true few-shot training setting, where $N=\left| Y \right|$ and $K$ is the number of samples per event type.
Then an instance $x_i$, randomly sampled from $\textbf{X}$, is fed into the next multi-step prompt module to obtain the trigger $t_i$, which is accordingly input into the event classifier together with $x_i$ to generate a $d$-dimension event embedding $\textbf{e}_0 \in \mathbb{R}^d$. 
Finally, the event embedding $\textbf{e}_0$ is mapped into the event vector space $\textbf{E} = \{\textbf{e}_1, \textbf{e}_2, \ldots, \textbf{e}_N\}$ in the prototypical network, where $\textbf{e}_i \in \mathbb{R}^d$ is a $d$-dimension vector of event type $y_i$.
Thus, we can get the probability $p_i$ of each event label, and the label with the largest probability is the predicted one.

\subsection{Under-sampling}
\label{Under-sampling}

\begin{figure}[t]
\centering
\begin{minipage}[ht]{0.49\columnwidth}
	\includegraphics[clip,trim=0mm 0mm 0mm 0mm,width=\columnwidth]{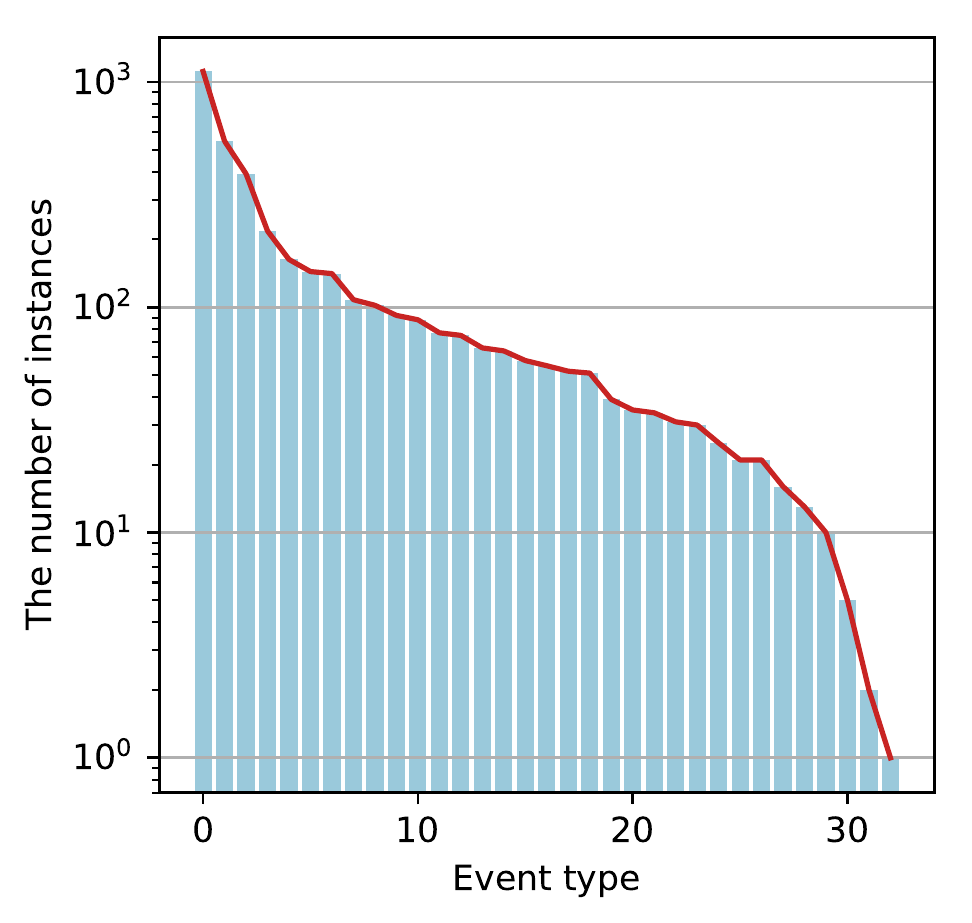}
	\subcaption{\emph{ACE-2005}}
	\label{Figure3.1}
\end{minipage}
\begin{minipage}[ht]{0.49\columnwidth}
	\includegraphics[clip,trim=0mm 0mm 0mm 0mm,width=\columnwidth]{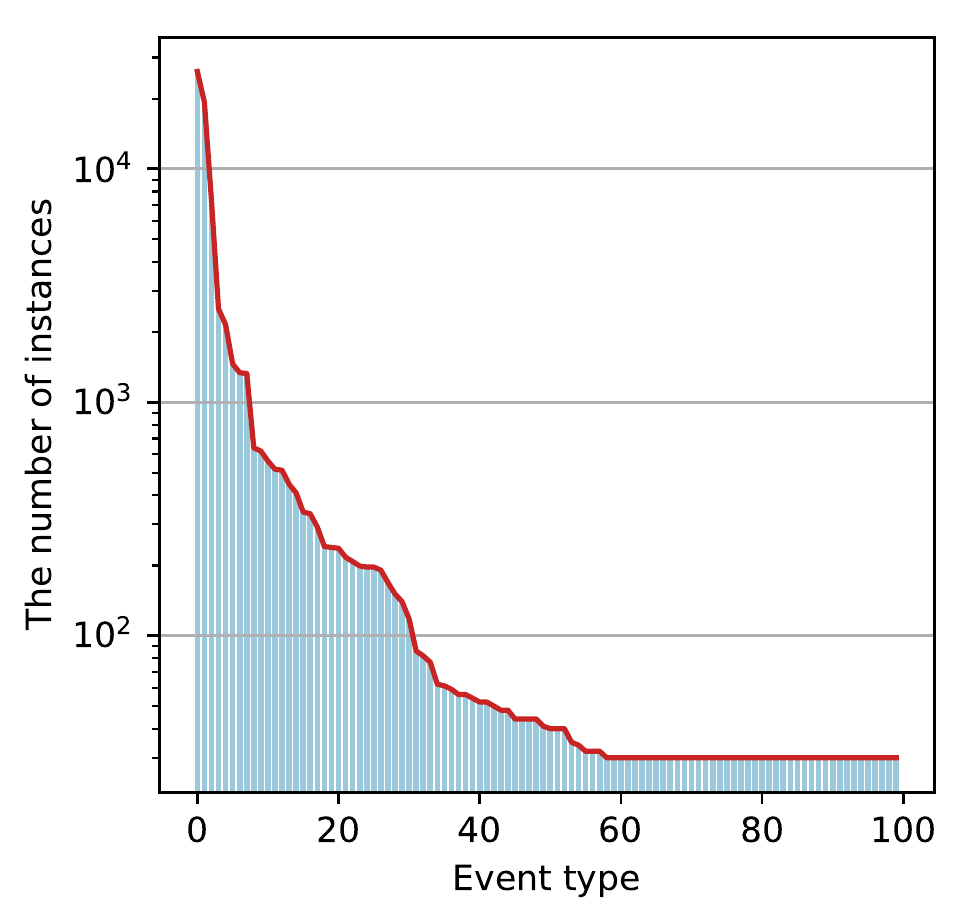}
	\subcaption{\emph{FewEvent}}
	\label{Figure3.2}
\end{minipage}
\caption{Distribution of the number of instances with different event types on \emph{ACE-2005} and \emph{FewEvent}. The numbers (y-axis) are exponentially distributed and the event types (x-axis) are ordered according to their corresponding frequency.}
\label{Figure3}
\end{figure}

As the frequency of labels in the event label set $Y$ is extremely unbalanced and obeys a long-tail distribution, shown in Fig.~\ref{Figure3}, it brings an unpredictable deviation to event detection.
To avoid such deviation and enhance the generalization ability of scarce event types, we utilize an under-sampling module into ED, which selects the same number of instances with each event type to form a novel training and validation set without label deviation.

\vspace{-1pt}
\begin{figure*}[t]
\centering
\includegraphics[width=1.0\textwidth]{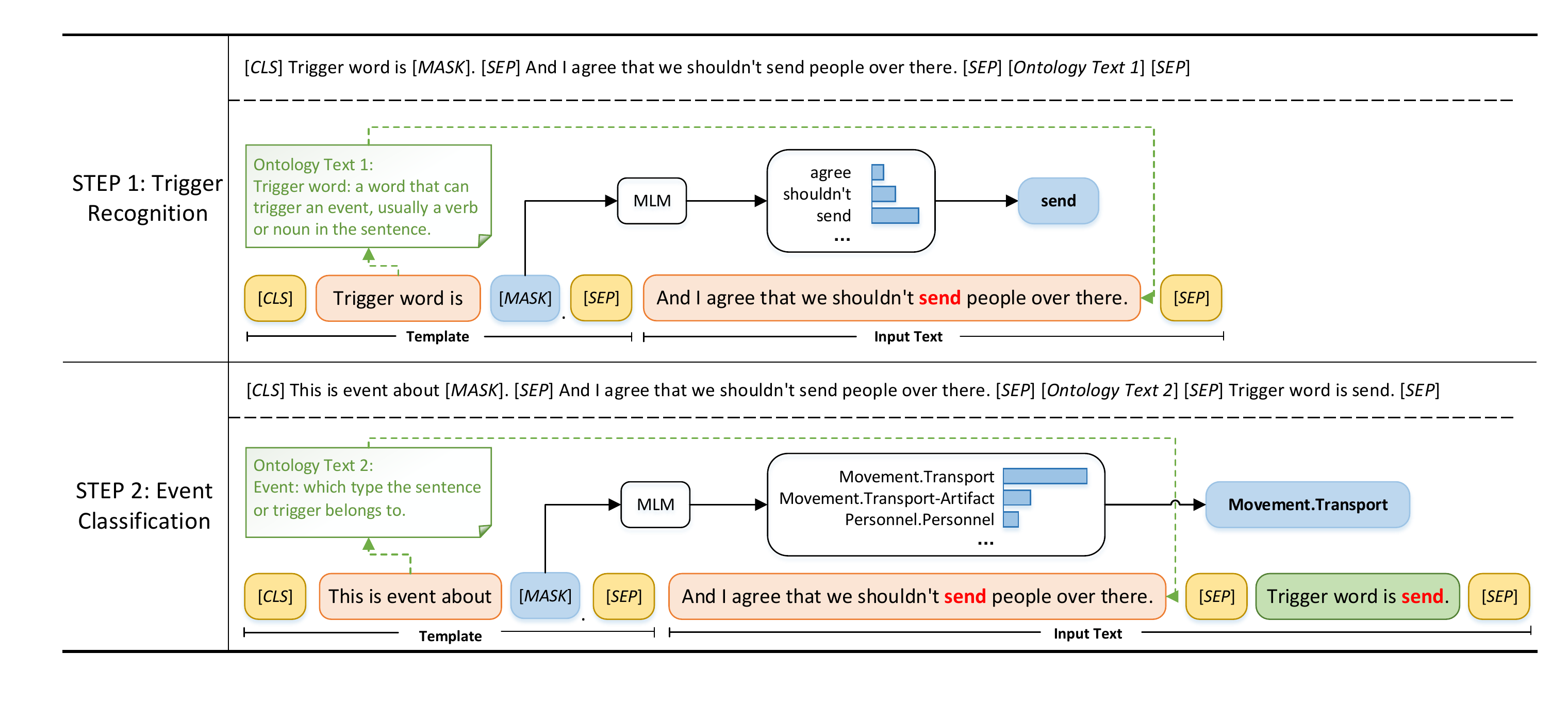}
\caption{The architecture of the multi-step prompt module.}
\label{figure4}
\end{figure*}

In detail, given an ED dataset, we randomly sample $K$ instances for each event type to form a $K$-shot training set:
\begin{equation}
\label{equation1}
\textbf{X}_{train} =  \left[ {\begin{array}{*{20}{c}}
{x_1^1}&{x_1^2}& \cdots &{x_1^K}\\
{x_2^1}&{x_2^2}& \cdots &{x_2^K}\\
\vdots & \vdots & \ddots & \vdots \\
{x_N^1}&{x_N^2}& \cdots &{x_N^K}
\end{array}} \right].
\end{equation}
After that, we repeat this operation on the rest of instances to generate a $K$-shot validation set $\textbf{X}_{valid}$ that does not intersect with $\textbf{X}_{train}$, i.e., $ \textbf{X}_{train} \cap \textbf{X}_{valid} = \emptyset$. 
Finally, the remaining unsampled instances consist the test set $\textbf{X}_{test} = \textbf{X} - \textbf{X}_{valid} - \textbf{X}_{train}$.

The newly constructed training, validation, and test sets can accommodate the true few-shot learning paradigm that meets the low resource scenarios in reality to effectively evaluate the performance of FSED models.

\subsection{Multi-step prompt}
\label{Multi-step Prompt}

Since the conventional ED paradigm $x \to (x, t) \to y$ is a multi-step process, we extend the general one-step prompt into the multi-step prompt, performing the two subtasks coherently in one iteration and synchronously training the two consecutive subprompts.
We depict how the multi-step prompt module works in Fig.~\ref{figure4}. 
It mainly contains two steps, i.e., trigger recognition (see Section~\ref{Trigger recognition}) and event classification (see Section~\ref{Event classification}).
In addition, a supplemental knowledge-enhanced ontology used in each step is discussed in Section~\ref{Knowledge-enhanced ontology}.

\subsubsection{Trigger recognition}
\label{Trigger recognition}
Identifying triggers can be regarded as a text annotation task.
First, we manually construct ``\emph{Trigger word is} [\emph{MASK}]." as the prefixed prompt template $\mathcal{T}_1$, where [\emph{MASK}] is the masked position to match the trigger word $t_i$ in an event mention $x_i$.
Then we concatenate the template $\mathcal{T}_1$ with each event mention $x$ (e.g., ``\emph{And I agree that we shouldn't send people over there.}" in Fig.~\ref{figure4}) to obtain the modified prompt $f_1$ as:
\begin{equation}
\label{equation2}
f_1(x)=[CLS] \underline{Trigger\ word\ is\ [MASK].} [SEP] x [SEP].
\end{equation}

In detail, given an event mention $x$, a sequence $\textbf{w} = ({w_1},{w_2}, \cdots ,{w_L})$ is obtained after word segmentation, where $L$ is the mention length.
The trigger $t$ can be represented by the embedding of [\emph{MASK}] that is filled via a masked language modeling process.
Then the trigger probability distribution is obtained by mapping  the vocabulary list of PLMs to $\textbf{w}$, denoted as $P_t$:
\begin{equation}
\label{equation3}
\begin{aligned}
P_t &= P\left( \left. [MASK] = t \right|f_1(x)\right)\\
&= P\left( \left. t = {w_j} \right|\textbf{w} \right).
\end{aligned}
\end{equation}
The candidate word $w_j \in \textbf{w}$ with the highest probability will be recognized as the trigger word.
Then the annotated sequence of the event mention is produced, i.e., $\textbf{w}^{\prime} = ({w_1},{w_2}, \cdots, t, \cdots ,{w_L})$.
The loss function of trigger recognition is defined as a cross-entropy loss $L_t$ as:
\begin{equation}
\label{equation4}
L_t =  - \sum_{j = 1}^L {{\hat t}_j\log \left(P\left( \left. t = {w_j} \right|\textbf{w} \right)\right)},
\end{equation}
where the gold trigger label is extend into a one-hot vector $ \hat t = \{\hat t_j\}_{j=1}^L$.

\subsubsection{Event classification}
\label{Event classification}
After obtaining the predicted trigger word, we can classify the mentions into a predefined event label, dubbed as event classification.
The prefixed prompt template $\mathcal{T}_2$ used here is ``\emph{This is event about} [\emph{MASK}].", where [\emph{MASK}] can be regarded as the event embedding $\textbf{e}_0 \in \mathbb{R}^d$ to represent the event context.
Next, for generating $\textbf{e}_0$, the new assembled prompt $f_2$ is fed into the \acf{MLM} as:
\begin{equation}
\label{equation5}
f_2(x')=[CLS] \underline{This\ is\ event\ about\ [MASK].} [SEP] x' [SEP],
\end{equation}
where $x'$ is the initial event mention $x$ integrated with the corresponding trigger word (e.g., ``\emph{And I agree that we shouldn't send people over there.} [\emph{SEP}] \emph{Trigger word is send.}" in Fig.~\ref{figure4}).
Note that the trigger word is obtained by the trigger recognizer at the validation or test stage, while it is the ground-truth trigger in the training stage.

Then, given a set of event labels $Y = \{y_1, y_2, \ldots, y_N\}$ and the generated event embedding $\textbf{e}_0$, the event probability distribution can be symbolized as $P_y$:
\begin{equation}
\label{equation6}
\begin{aligned}
P_y &= P\left( \left. [MASK] = y \right|f_2(x')\right)\\
&= P\left( {\left. y = {y_j} \right|Y} \right)\\
&= \left( p_j \right)_N,
\end{aligned}
\end{equation}
where $p_j (j = 1,2, \cdots ,N)$ is the predicting probability of event label $y_j$, and $\arg\max\limits_{y_j} {p_j}$ is the target event label. To train the event classifier, we employ a cross-entropy loss as the optimization objective:
\begin{equation}
\label{equation7}
L_y =  - \sum_{j = 1}^N {{\hat y}_j\log \left( p_j \right)},
\end{equation}
where the actual event label is represented as a one-hot vector $\hat y = \left( {\hat y}_1,{\hat y}_2, \cdots ,{\hat y}_N \right)$.
To sum up, the total loss function of MsPrompt can be expressed as:
\begin{equation}
\label{equation8}
L= \alpha L_t + \beta L_y,
\end{equation}
where $\alpha, \beta \in \mathbb{R}$ are the adjustable parameters of the trigger recognizer and the event classifier, respectively.

\subsubsection{Knowledge-enhanced ontology}
\label{Knowledge-enhanced ontology}
In the trigger recognizer and the event classifier, we employ the distinct prompt template $\mathcal{T}_1$ and $ \mathcal{T}_2$ as clues to detect the target trigger and the event type by the process of masked language modeling.
However, the ``\emph{trigger word}" and ``\emph{event}" are still hard to understand for the PLMs.
Therefore, we introduce the knowledge-enhanced ontology to extend the semantics of these key words to well elicit the latent knowledge from PLMs and connect with the current task.

In particular, for the trigger recognizer, we add an ontology text $\mathcal{O}_1$ ``\emph{Trigger word: a word that can trigger an event, usually a verb or noun in the sentence.}" after the event mention to further elaborate the meaning of ``\emph{trigger word}".
Analogously, in the event classifier, another ontology text $\mathcal{O}_2$ ``\emph{Event: which type the sentence or trigger belongs to.}" is placed between the mention and the trigger word, which helps the prompt model clarify the objective of event classification.

\subsection{Prototypical network}
\label{Prototypical network}
Due to the disabled generalization caused by the severe trigger bias, classifying scarce labels is challenging in the true few-shot setting.
Therefore, when generating the probability distribution of the predicted event label in Section~\ref{Event classification}, we abandon the convention of applying a verbalizer to map the embedding $\textbf{e}_0$ from the vocabularies of PLMs to the label space $Y$ simply.
Instead, we introduce a prototypical network that clusters all instances to obtain the centroid of each cluster as the representation of the event labels.
With a strong generalization ability, this module can reduce the interference of edge instances in dense clusters, i.e., the noise of high-frequency labels, and can also effectively capture the inter-class relationship as well as the spatial information of all samples in the sparse clusters to boost the discrimination ability of the low-frequency labels.

In particular, given an event embedding $\textbf{e}_0$ obtained from the event classifier and the randomly initialized event vector space $\textbf{E} = \{\textbf{e}_1, \textbf{e}_2, \ldots, \textbf{e}_N\}$, we gauge the distance between $\textbf{e}_0$ and $\textbf{e}_i \in \textbf{E} $ by Euclidean Metric.
The cluster centroid is further calculated as the label representation $\textbf{e}_i \in \mathbb{R}^d $, all of which form a prototypical network $\textbf{E}$.
Then, the prototypical network is updated synchronously with the prompt model by the optimization objective.
Thus, the predicted probability $p_j \in \mathbb{R}$ of label $y_j$ can be expressed as:
\begin{equation}
\label{equation9}
p_j =\frac{\exp \left(- \mathcal{D} \left( \textbf{e}_0, \textbf{e}_j\right)\right)}{\sum_{n=1}^N \exp \left(-\mathcal{D}\left(\textbf{e}_0, \textbf{e}_n\right)\right)},
\end{equation}
where $\mathcal{D}(\cdot, \cdot)$ returns the Euclidean distance.
Then we obtain the probability distribution $P_y$ in Eq.(\ref{equation6}) to identify the predicted event label with the maximal probability, 
which can be regarded as the centroid closest to $\textbf{e}_0$ in the prototypical network.

\section{Experiments}
\label{Experiments}

In this section, we first introduce two widely-used datasets for event detection in Section~\ref{Datasets}.
Then we detail the research questions and experimental configurations in Section~\ref{Research questions and configurations}.
Finally, we list several baselines in Section~\ref{Model summary}.

\subsection{Datasets}
\label{Datasets}
We evaluate the performance of MsPrompt and the baselines on two datasets, i.e., \emph{ACE-2005} \footnote{https://catalog.ldc.upenn.edu/LDC2006T06} and \emph{FewEvent} \footnote{https://github.com/231sm/Low\_Resource\_KBP}.
We follow \citet{naacl21/Zhang} to check whether the trigger words are consistent with the annotated index range and delete the inconsistent samples.
After that, the statistics of \emph{ACE-2005} and \emph{FewEvent} are provided in Table~\ref{Table1}.
Since the average trigger length of both datasets is close to 1, we set the trigger recognizer to select one word per input.
\begin{table}[t]
\centering
\caption{\textbf{Statistics of \emph{ACE-2005} and \emph{FewEvent} used in our experiments.}}
\setlength{\tabcolsep}{0.01\linewidth}{
	\begin{tabular}{l lll}
		\toprule
		Statistics & {\emph{ACE-2005}} & {\emph{FewEvent}} \\
		\midrule
		\#  Event types                          &33     &100\\
		\#  Event mentions                       &3,653  &68,694\\
		\#  Average number of event mentions     &110.70 &686.94\\
		\#  Average mention length               &27.31  &32.78\\
		\#  Average trigger length               &1.23   &1.01\\
		\bottomrule
\end{tabular}}
\label{Table1}
\end{table}

As shown in Table~\ref{Table2}, to evaluate the FSED performance of our proposal and the baselines, we adopt the true few-shot training settings \citep{ACL21/Gao} with different sample size $K\in\{4,8,16,32\}$ to obtain the $K$-shot training set $\textbf{X}_{train}$ and validation set $\textbf{X}_{valid}$.
The test set is formed with the remaining instances.
It is worth noting that we ignore the event types that the number of mentions is less than or equal to $2K$ and can not be divided into the train/valid/test sets under our true few-shot setting.
On a general full-data set, we divide the train/valid/test sets at a ratio of 8:1:1.

Since $\textbf{X}_{test}$ is still perturbed by the trigger bias, the most impartial evaluation should be based on an unbiased test set.
Therefore, following \citet{cikm21/Wang}, we employ three sampling methods to construct a novel test set as follows:
\begin{enumerate}[leftmargin=*,nosep]
\item \textbf{Instance Uniform Sampling (IUS)} selects $K$ mentions from each event type randomly.

\item \textbf{Trigger Uniform Sampling (TUS)} samples $K$ mentions uniformly from each trigger of one event type .

\item \textbf{COnfusion Sampling (COS)} picks $K$ mentions uniformly from confusing triggers, i.e., similar to triggers of other event types, of one event type.
\end{enumerate}
Compared with the biased full-test set $\textbf{X}_{test}$ and the dataset after IUS, the datasets under the aforementioned sampling methods TUS and COS, are stripped of the trigger bias that leads to exaggerated performance.

\begin{table*}[t]
\centering
\caption{Size of the train/valid/test sets under $K$-shot and full-data setting.}
\setlength{\tabcolsep}{0.005\linewidth}{
	\begin{tabular}{l cc cc cc cc cc cc cc cc cc cc }
		\toprule
		\multirow{2.5}{*}{Statistics}&
		\multicolumn{5}{c}{\emph{ACE-2005}}&
		\multicolumn{5}{c}{\emph{FewEvent}}\\
		\cmidrule(lr){2-6} \cmidrule(lr){7-11}
		&4-shot&8-shot&16-shot&32-shot&Full-data &4-shot&8-shot&16-shot&32-shot&Full-data \\
		\midrule
		\# Event type &30 &27 &21 &13 &33  &100 &100 &56 &34 &100\\
		\# Train instances   &120	&216 &336 &416 &2,921  &400 &800	&896 &1,088 &54,954\\
		\# Valid instances   &120	&216 &336 &416 &366  &400 &800 &896 &1,088 &6,870\\
		\# Test instances    &3,405&3,176&2,791&2,232&366  &67,894&67,094&65,692&64,323&6,870\\
		\bottomrule
\end{tabular}}
\label{Table2}
\end{table*}

\subsection{Research questions and configurations}
\label{Research questions and configurations}

\subsubsection{Research questions}
\label{Research questions}
To evaluate the performance of MsPrompt, we focus on the following research questions:
\begin{enumerate}[label=(RQ\arabic*),leftmargin=*,nosep]
\item Does our model MsPrompt improve the performance for the true few-shot event detection compared to the state-of-the-art baselines?
\item Can our model perform better than the best baseline in strict low-resource scenarios?
\item How is the impact of sampling methods on the model performance, i.e., IUS, TUS, and COS?
\item How is the performance of MsPrompt affected by the input length?
\item How is the performance of MsPrompt affected by the input sequence?
\item Which part of the model has the greatest contribution to the task?
\item Does our model solve the issues caused by the trigger bias?
\end{enumerate}

\subsubsection{Model configurations}
\label{Model configurations}
We make an under-sampling operation at a fixed random seed 42, and use the \emph{bert-base-uncased} \footnote{https://huggingface.co/bert-base-uncased/tree/main} to obtain the representation of each event mention.
The parameters $\alpha$ and $\beta$ in Eq.(\ref{equation8}) are set to 1.
The batch size is set to 32 and 128 at the training and test stages respectively. In addition, AdamW \citep{iclr19/Loshchilov} is utilized as the model optimizer.
For \emph{ACE-2005} and \emph{FewEvent}, the epoch number is 500, 100, the learning rate of \emph{bert-base-uncased} is $1e^{-6}$, $1e^{-5}$, and the learning rate of other part is $1e^{-3}$, $1e^{-2}$, respectively.
Moreover, when the model loss has no more reduction after 1,000 iterations, we terminate the training process according to the early stop strategy.
We implement MsPrompt under several random seeds to average the results, containing accuracy, weighted precision, weighted recall, and weighted F1-score.
For brevity, we omit ``\emph{weighted}" in the following tables and figures.
In addition, the metric used in Table~\ref{Table3},~\ref{Table4},~\ref{Table5} is weighted F1-score.

\subsection{Model summary}
\label{Model summary}
The following models are discussed.
The first group of baselines is based on different metric learning methods as follows.
\begin{itemize}[leftmargin=*,nosep]
\item\textbf{Neural-Bert} applies a traditional neural network to map the output embedding from the hidden dim to the event label space directly, and thus obtain the label probability distribution of the embedding. 
\item\textbf{Relation-Bert} \citep{cvpr18/Sung} follows the idea of the relation network to perform event detection. The embedding module and the relation module used here are the prototypical network and a single fully connected layer respectively.
\item\textbf{Proto-Bert} \citep{nips17/Snell} employs the prototypical network as the event classifier to calculate the Euclidean distance between the event embedding and each event type. The nearest event type is then the prediction label.
\end{itemize}
The second group of baselines is prompt-based methods as follows.
\begin{itemize}[leftmargin=*,nosep]
\item\textbf{KiPT} \citep{coling22/Li} utilizes \emph{T5} as the event encoder to obtain the soft prompt template, and introduces the external knowledge bases to construct the knowledge-injected prompts.
\item\textbf{ZEOP} \citep{naacl21/Zhang} is based on prompt learning and ordered contrastive learning. Here, \emph{bert-base-uncased} is applied as the encoder. It applies a one-step prompt to obtain the trigger token and event embedding.
\end{itemize}

\section{Results and Discussion}
\label{Results and Discussion}
First, we discuss the overall FSED performance of our proposal and the baselines in Section~\ref{Overall performance}, and then explore their performance under the strict low-resource scenarios in Section~\ref{Strict low-resource performance}.
Next, we conduct the comprehensive experiments to evaluate the performance of MsPrompt under different sampling methods (Section~\ref{Debiasing Performance}), input length (Section~\ref{Impact of Input length}), and input sequence (Section~\ref{Impact of Input sequence}).
After that, we perform an ablation study to explore the effect of each part in our proposal in Section~\ref{Ablation study}.
Finally, we conduct a case study to verify the contribution of our model in mitigating the trigger bias in Section~\ref{Case study}.

\subsection{Overall performance}
\label{Overall performance}

\begin{table*}[t]
\centering
\caption{Overall performance. The results of the best-performing baseline and the best performer in each column are underlined and boldfaced, respectively.}
\setlength{\tabcolsep}{0.02\linewidth}{
	\begin{tabular}{l cc cc cc cc cc cc cc cc}
		\toprule
		\multirow{2.5}{*}{Model}&
		\multicolumn{4}{c}{\emph{ACE-2005}}&
		\multicolumn{4}{c}{\emph{FewEvent}}\\
		\cmidrule(lr){2-5} \cmidrule(lr){6-9} 
		&4-shot&8-shot&16-shot&32-shot &4-shot&8-shot&16-shot&32-shot\\
		\midrule
		Neural-Bert   &18.43 &21.54 &42.86 &46.65 &13.87 &27.11 &55.69 &73.32\\
		Relation-Bert &44.09 &47.95 &52.17 &60.33 &41.53 &44.30 &56.03 &56.84\\
		Proto-Bert    &15.20 &31.24 &43.24 &57.13 &43.09 &58.61 &70.07 &72.95\\
		\midrule
		KiPT &\underline{45.50} &50.10 &54.20 &56.40 &54.87 &61.34 &65.23 &68.72\\
		ZEOP &44.52 &\underline{58.92} &\underline{\textbf{62.99}} &\underline{69.66} &\underline{57.63} &\underline{\textbf{65.71}} &\underline{71.41} &\underline{73.79}\\
		\midrule
		\textbf{MsPrompt} &\textbf{46.30} &\textbf{60.29} &61.88 &\textbf{69.94} &\textbf{60.67} &60.83 &\textbf{73.35} &\textbf{73.90}\\
		\bottomrule
\end{tabular}}
\label{Table3}
\end{table*}

For answering RQ1, we present the event detection performance of our proposal and the baselines on two public datasets: \emph{ACE-2005} and \emph{FewEvent}.
Following \citep{coling22/Li}, we evaluate the models under a few-shot setting with $K \in \{4, 8, 16, 32\}$ in Table~\ref{Table3}.

For the baselines, in general, shown in Table~\ref{Table2}, as the number of instances increases, the performance of mentioned models consistently goes up in the low-resource scenarios.
In detail, we observe that most models perform better on \emph{FewEvent} than on \emph{ACE-2005}.
It can be explained that \emph{FewEvent} contains more train samples than \emph{ACE-2005} to help the models classify the event type correctly.
In addition, the prompt-based learning models (e.g., ZEOP and MsPrompt) generally outperform the three traditional baselines based on metric learning, which confirms the high applicability of prompt learning to few-shot scenarios.
Among three metric learning models, Proto-Bert achieves a better performance on \emph{FewEvent} than Neural-Bert and Relation-Bert generally.
However,  Relation-Bert outperforms the other two models on \emph{ACE-2005}, and performs relatively stable under various $K$-shot settings on both datasets.
This is due to the insensitivity of Relation-Bert to the label space of the datasets and the size of the training set.

Next, for the prompt-based models, MsPrompt achieves a notable improvement against the baselines for most cases.
For instance, 
MsPrompt performs the best for the cases under $K = 4, 8, 32$ on \emph{ACE-2005} and $K = 4, 16, 32$ on \emph{FewEvent}.
Instead, ZEOP performs well on \emph{ACE-2005} with $K = 16$ and on \emph{FewEvent} with $K = 8$,
which can be attributed to the high similarity of the inter-class samples under such settings.
Besides, ZEOP obtains more supervised signals than our proposal by introducing additional contrastive samples for training.
Nevertheless, MsPrompt is implemented in the true few-shot training setting without any supplementary samples.
In particular, the improvements of MsPrompt over the best-performing baseline on \emph{ACE-2005} under the 4-shot, 8-shot, and 32-shot settings are 0.80\%, 1.37\%, and 0.28\%, respectively, and 3.04\%, 1.94\%, and 0.11\% on \emph{FewEvent} with 4-shot, 16-shot, 32-shot, respectively.

\subsection{Strict low-resource performance}
\label{Strict low-resource performance}

\begin{table}[t]
\centering
\caption{The performance of MsPrompt and ZEOP in the strict low resource scenarios under the $K$-shot setting, where $K \in \{1,2,3,4\}$.}
\setlength{\tabcolsep}{0.02\linewidth}{
	\begin{tabular}{l l cc cc cc cc}
		\toprule
		Dataset &Model &1-shot &2-shot &3-shot &4-shot\\
		\midrule
		\multirow{2}{*}{\emph{ACE-2005}} &MsPrompt &24.16 &32.66 &43.10 &46.30\\
		&ZEOP &17.22 &26.73 &37.02 &44.52\\
		\midrule
		\multirow{2}{*}{\emph{FewEvent}} &MsPrompt &29.53 &41.49 &51.75 &60.67\\
		&ZEOP &28.14 &36.29 &40.32 &57.63\\
		\bottomrule
\end{tabular}}
\label{Table4}
\end{table}
To answer RQ2, we conduct an extensive experiment to evaluate the performance of MsPrompt and the best-performing baseline ZEOP in the strict low-resource scenarios, as shown in Table~\ref{Table4}.

Clearly, MsPrompt consistently outperforms ZEOP in the few-shot settings with $K = 1,2,3,4$ on both datasets.
For instance, on \emph{ACE-2005}, MsPrompt presents 6.94\%, 5.93\%, 6.08\%, 1.78\% improvements in terms of weighted F1-score against ZEOP under the 1-shot, 2-shot, 3-shot, 4-shot setting, respectively. For \emph{FewEvent}, the corresponding improvements are 1.39\%, 5.20\%, 11.43\% and 3.04\%.
It can be explained by the outstanding performance of prompt learning in the strict low-resource scenarios, i.e., exploring the potential knowledge from PLMs to enhance the model training.
In contrast with ZEOP, MsPrompt employs the whole process prompt model for two consecutive sub-tasks in event detection, which maximizes the advantages of prompt learning in few-shot learning, and makes full use of the latent knowledge in PLMs.

As shown in Table~\ref{Table4}, when $K$ increases, the performance of MsPrompt and ZEOP improves on both datasets.
A similar pattern of results still exists in this experiment, i.e., both MsPrompt and ZEOP achieve a higher performance in terms of weighted F1-score on \emph{FewEvent} than that on \emph{ACE-2005}.
Additionally, we find that an obvious performance improvement of MsPrompt can be observed when the number of training samples achieves a moderate point. 
For instance, when switching from 2-shot to 3-shot on \emph{ACE-2005}, MsPrompt presents the largest improvement gap, i.e., 10.44\% from 32.66\% to 43.10\%.
This implies that increasing the number of instances can indeed accelerate the rate of sample utilization and thus enhance the model performance.

\subsection{Debiasing Performance}
\label{Debiasing Performance}

\begin{figure}[t]
\centering
\begin{minipage}[ht]{0.49\columnwidth}
	\includegraphics[clip,trim=0mm 0mm 0mm 0mm,width=\columnwidth]{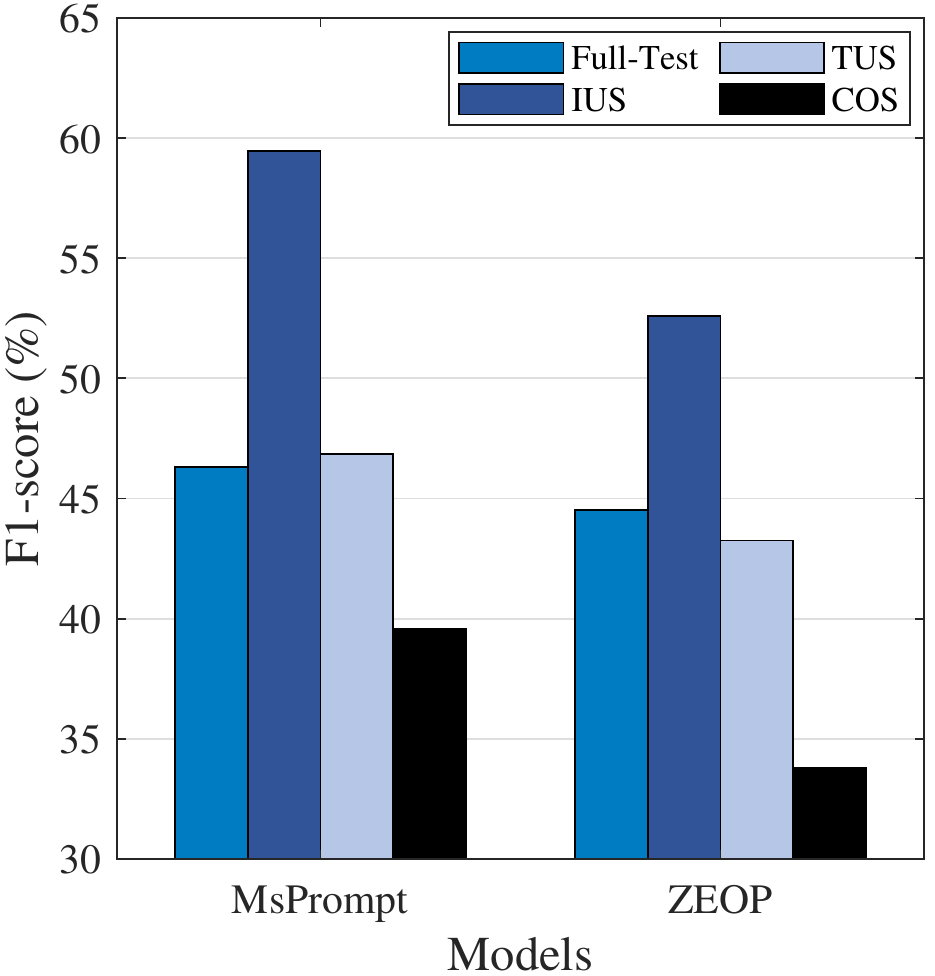}
	\subcaption{\emph{ACE-2005}}
	\label{Figure5.1}
\end{minipage}
\begin{minipage}[ht]{0.49\columnwidth}
	\includegraphics[clip,trim=0mm 0mm 0mm 0mm,width=\columnwidth]{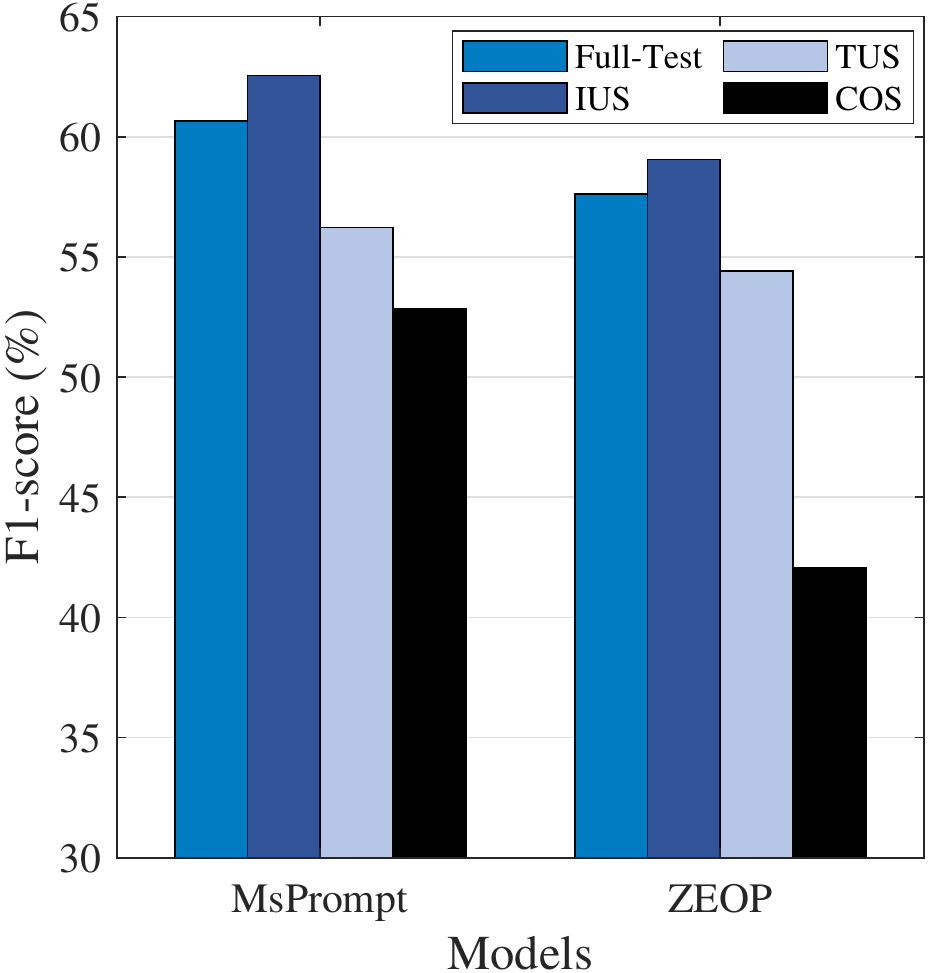}
	\subcaption{\emph{FewEvent}}
	\label{Figure5.2}
\end{minipage}
\caption{Model performance on \emph{ACE-2005} and \emph{FewEvent} under the 4-shot setting on the full or sampled test set. For IUS, TUS, and COS, we sample 4 mentions from each event type.}
\label{Figure5}
\end{figure}

To answer RQ3, we evaluate the debiasing performance of MsPrompt and the best-performing baseline ZEOP under the IUS, TUS, COS, and initial full-test set $\textbf{X}_{test}$ (dubbed as Full-Test) in Fig.~\ref{Figure5}.

As shown in Fig.~\ref{Figure5}, for both models, we observe that the results on IUS, i.e., the event-uniform test set, are better than that on Full-Test with a long-tail distribution of event types.
It may be deduced that the datasets where the labels exhibit a long-tailed distribution limit the few-shot event detection performance.
Compared with ZEOP, our  MsPrompt model not only performs well on the unbalanced datasets but also on the event-uniform datasets, which reflects an outstanding robustness.

In detail, for the unbiased test set, the results of TUS and COS on \emph{ACE-2005} and \emph{FewEvent} have different degrees of decline compared with the results of IUS and Full-Test.
This indicates that the trigger bias in datasets makes the event detection model highly rely on such trigger clues, and the actual FSED performance is overestimated.
Moreover, MsPrompt continues to outperform ZEOP on TUS and COS, which reveals that MsPrompt still has a robust and outstanding performance improvement in a fair debiasing scenario.
This improvement is more intuitive on COS than TUS, presenting about 5\% and 10\% on \emph{ACE-2005} and \emph{FewEvent} respectively, indicating a better prediction for confusing triggers of MsPrompt than ZEOP.
This is due to that MsPrompt focuses on the context rather than on misclassifying the confusing trigger words by trusting the trigger bias.

In addition, compared with the results on Full-Test, the results of MsPrompt on TUS are fluctuated by no more than 5\% on both datasets, and show a slight improvement on \emph{ACE-2005}, proving that MsPrompt has an outstanding performance for debiasing few-shot event detection.
It demonstrates that MsPrompt can leverage the supplementary ontology text and prompt learning to model the whole event mentions and fully mine the prior knowledge in PLMs, which effectively overcomes the problems of context-bypassing and disabled generalization caused by the trigger bias.

\subsection{Impact of Input length}
\label{Impact of Input length}

\begin{figure}[t]
\centering
\begin{minipage}[ht]{0.49\columnwidth}
	\includegraphics[clip,trim=0mm 0mm 0mm 0mm,width=\columnwidth]{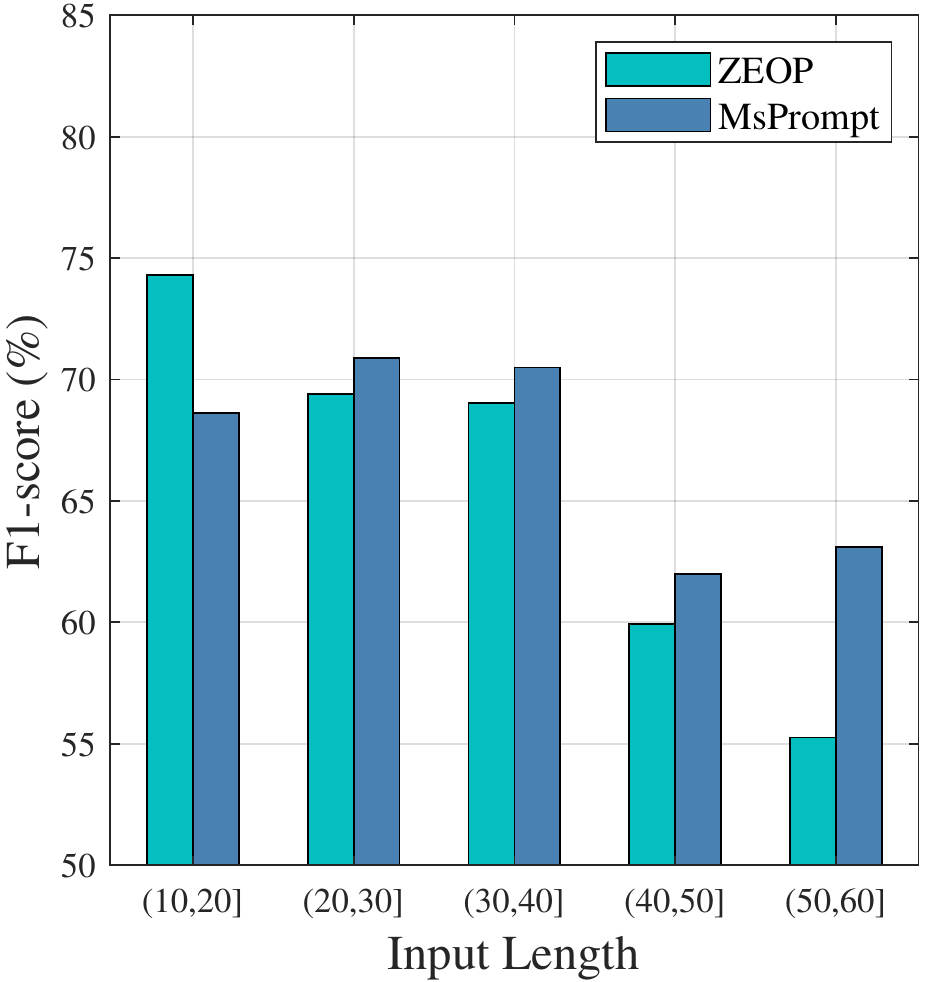}
	\subcaption{\emph{ACE-2005}}
	\label{Figure6.1}
\end{minipage}
\begin{minipage}[ht]{0.49\columnwidth}
	\includegraphics[clip,trim=0mm 0mm 0mm 0mm,width=\columnwidth]{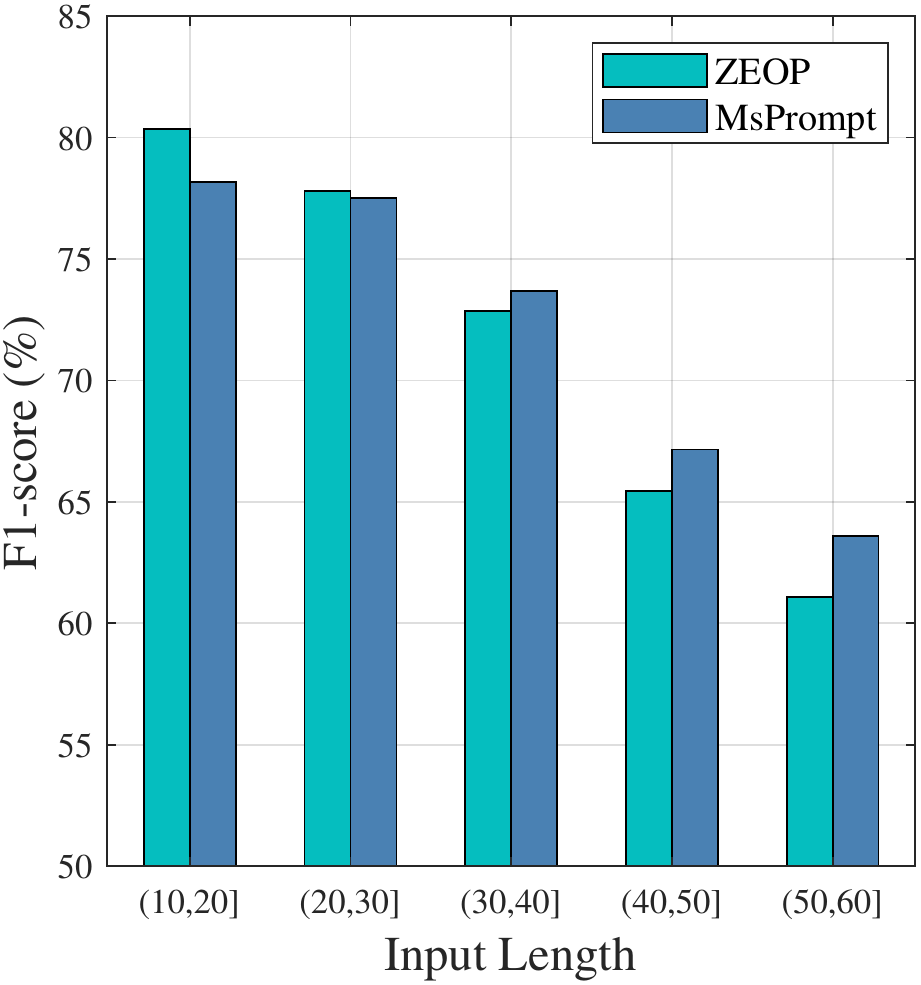}
	\subcaption{\emph{FewEvent}}
	\label{Figure6.2}
\end{minipage}
\caption{Impact of input length on \emph{ACE-2005} and \emph{FewEvent} under the 32-shot setting.}
\label{Figure6}
\end{figure}

In order to discuss the impact of input length in RQ4, we compare the performance of MsPrompt and the best-performing baseline ZEOP under a variety of length intervals on \emph{ACE-2005} and \emph{FewEvent} under the 32-shot setting.
Since the average mention length in Table~\ref{Table1} is 27.31 on \emph{ACE-2005} and 32.78 on \emph{FewEvent}, we select five equally spaced length intervals, i.e., (10,20], (20,30], (30,40], (40,50], and (50,60].
For example, the test instances in ``(10,20]" are composed of samples with the mention length between 10 and 20.
Thus, we obtain five segmented test sets from the corresponding test set of 32-shot on \emph{ACE-2005} and \emph{FewEvent}, respectively.
The results are shown in Fig.~\ref{Figure6}. 

We find that an overall trend is that the longer the input length, the lower the model performance of MsPrompt and ZEOP present on both datasets.
It may be attributed to the fact that long sentences are more complicated and the semantics are more difficult to understand.
And with the increase of mention length, more noise can be carried in the input and thus the challenge to the trigger recognizer will rise, which will affect the performance of the few-shot event detection.
This decline is more obvious on \emph{FewEvent}, where  more indistinguishable event labels are included.

In addition, on \emph{ACE-2005}, compared with ZEOP, 
MsPrompt first loses the competition in the group of (10, 20] and then overtakes it in the groups of (20,30], (30,40], (40,50] and (50,60].  In addition,  its advantages become more prominent as the length increases.
This trend also exists on \emph{FewEvent} and MsPrompt outperforms ZEOP from the length interval of (30, 40].
Therefore, we can conclude that MsPrompt has an advantage against ZEOP on long inputs.
Furthermore, as the length and complexity of sentence increase, MsPrompt has more obvious advantages in few-shot event detection than ZEOP.
This may benefit from the excellent semantic modeling ability of the MLM.
Even in the long event mentions, the trigger recognizer can identify the key information and trigger words  accurately.
In addition,  the event classifier can also combine the long mention, ontology text, and the identified trigger information together effectively improve the event detection performance.
In particular,  the supplement of ontology text enables the model to understand the current event detection task easily, such as understanding the definition of trigger word and event.
It helps the model to effectively avoid the interference of some invalid information of long mentions, and better seize the key information of the mentions.

\subsection{Impact of Input sequence}
\label{Impact of Input sequence}

Next, for answering RQ5, we evaluate the performance of MsPrompt under all combinations of input sequences in the trigger recognizer and event classifier. We present the results  in Table~\ref{Table6}.

\begin{table}[t]
\centering
\fontsize{9.5}{12}\selectfont
\caption{Impact of input sequence on \emph{ACE-2005} under the 32-shot setting. ``M", ``O", and ``T" means the event mention, ontology text, and trigger word, respectively. The results of MsPrompt are boldfaced. $^{\blacktriangledown}$ denotes the largest decline of performance in each column compared with MsPrompt.}
\setlength{\tabcolsep}{0.005\linewidth}{
	\begin{tabular}{l cc cc cc cc}
		\toprule
		\multirow{2.5}{*}{Sequence}&
		\multicolumn{1}{c}{Trigger recognition}&
		\multicolumn{3}{c}{Event detection}\\
		\cmidrule(lr){2-5}
		&Accuracy &Precision &Recall &F1-score\\
		\midrule
		\multicolumn{5}{c}{Trigger recognizer}\\
		\midrule
		\textbf{\# M + O} &\textbf{66.80} &\textbf{73.00} &\textbf{72.09} &\textbf{69.94}\\
		\# O + M &\enspace58.33$^{\blacktriangledown}$ &70.52 &69.44 &66.77\\
		\midrule
		\multicolumn{5}{c}{Event classifier}\\
		\midrule
		\textbf{\# M + O + T} &\textbf{66.80} &\textbf{73.00} &\textbf{72.09} &\textbf{69.94}\\
		\# M + T + O &66.85 &65.19 &62.19 &57.88\\
		\# O + M + T &66.89 &69.46 &68.10 &65.69\\
		\# O + T + M &65.91 &54.45 &22.49 &\enspace17.25$^{\blacktriangledown}$\\
		\# T + M + O &66.08 &54.87 &35.35 &28.83\\
		\# T + O + M &66.04 &\enspace54.39$^{\blacktriangledown}$ &\enspace22.31$^{\blacktriangledown}$ &17.29\\
		\bottomrule
\end{tabular}}
\label{Table6}
\end{table}

For the trigger recognizer, when the order of event mention and ontology text is changed to ``O + M", the accuracy of trigger recognition decreases by 8.47\%, which drops more than that of other sequence combinations in the event classifier.
Notably, the performance of event detection also decreases obviously.
For instance, the result of the weighted F1-score is reduced by 3.17\% compared to the default sequential combination in MsPrompt, i.e., ``M + O".

For the event classifier, the different sequence combinations of event mention, ontology text, and trigger word have little effect on the accuracy of trigger recognition with a small fluctuation around 1\%.
This slight fluctuation comes from the joint optimization of trigger loss and event loss in Eq.~\ref{equation9}.
However, for event detection, changing the order causes an obvious performance decline in terms of weighted F1-score.
For instance, compared with the default sequence of the event classifier in MsPrompt, i.e., ``M + O + T", the weighted F1-score of other sequence combinations in Table~\ref{Table6} decreased by 12.06\%, 4.25\%, 52.69\%, 41.11\%, and 52.65\%, respectively.
Among them, the combination of ``O + T + M" returns the greatest decline.

It is worth noting that the performance of both the ``O + T + M" and ``T + O + M" combinations decrease by more than 50\% when the event mention is placed at the end of the input.
According to the Recency Bias proposed in \citep{icml21/Zhao}, the prompt models have the tendency to obtain the information in the text closest to the prompt template.
Therefore, it can be deduced that the event mention plays a dominant role for the event detection performance, while the trigger word and ontology text play a relatively complementary role in the comparison.
This phenomenon suggests that we need to pay particular attention to the overall semantic information in the event mention for enhancing the performance of event detection, rather than using the simple trigger recognition and classification to cover the whole event information included in the sentence.

\subsection{Ablation study}
\label{Ablation study}

\begin{table}[t]
\centering
\fontsize{8.5}{12}\selectfont
\caption{Ablation study on \emph{ACE-2005} under the 32-shot setting. ``-" means removing the module from our proposal MsPrompt. The results of MsPrompt are boldfaced. $^{\blacktriangledown}$ denotes the largest decline of performance in each column compared with MsPrompt.}
\setlength{\tabcolsep}{0.005\linewidth}{
	\begin{tabular}{l cc cc cc cc}
		\toprule
		\multirow{2.5}{*}{Model}&
		\multicolumn{1}{c}{Trigger recognition}&
		\multicolumn{3}{c}{Event detection}\\
		\cmidrule(lr){2-5}
		&Accuracy &Precision &Recall &F1-score\\
		\midrule
		\textbf{MsPrompt} &\textbf{66.80} &\textbf{73.00} &\textbf{72.09} &\textbf{69.94}\\
		- Trigger recognizer &\enspace65.32$^{\blacktriangledown}$ &68.80 &67.20 &64.28\\
		- Event classifier   &66.76 &67.25 &66.31 &63.17\\
		- Ontology text      &66.58 &\enspace66.13$^{\blacktriangledown}$ &\enspace63.53$^{\blacktriangledown}$ &\enspace59.77$^{\blacktriangledown}$\\
		\bottomrule
\end{tabular}}
\label{Table7}
\end{table}

For RQ6, to check the contribution of different modules in MsPrompt to the event detection performance, we perform an ablation study using our proposal under the 32-shot setting of \emph{ACE-2005}.
In the ablation study, we separately remove three specific modules to explore their effects on MsPrompt, namely ``- Trigger recognizer", ``- Event classifier", and ``- Ontology text".
Among them, ``- Trigger recognizer" and ``- Event classifier" imply ignoring the trigger recognizer and the event classifier in the multi-step prompt model in Fig.~\ref{figure4}.
Correspondingly, we use [\emph{CLS}] to semantically model the raw event mentions directly and obtain the predicted trigger words as well as the event types.
In addition, ``- Ontology text" means to delete the implanted ontology text in both trigger recognizer and event classifier.
The ablation results are shown in Table~\ref{Table7}.

As shown in Table~\ref{Table7}, when the trigger recognizer is removed, the trigger recognition performance of the whole model decreases most severely, with a 1.48\% drop in accuracy from 66.80\% to 65.32\%.
This decline confirms the effectiveness of the trigger recognizer and its indispensability to our proposal MsPrompt.
For the event classifier, the removal of this module causes a sharp drop in terms of the weighted F1-score, which dropped by 6.77\%.
It is obvious that the event classifier plays a prominent role in boosting the performance of few-shot event detection.
When turning to ``- Ontology text", we observe that all metrics of event detection show the greatest decline. 
That is, the weighted precision, recall, and F1-score are decreased by 6.87\%, 8.56\%, and 10.17\%, respectively.
This fully demonstrates the outstanding contribution of ontology text to few-shot event detection, driving the prompt model to quickly learn the goal of event detection tasks in low-resource scenarios and truly guiding the training of PLMs with human prior knowledge.

\subsection{Case study}
\label{Case study}

\begin{figure}[t]
\centering
\includegraphics[width=0.8\columnwidth]{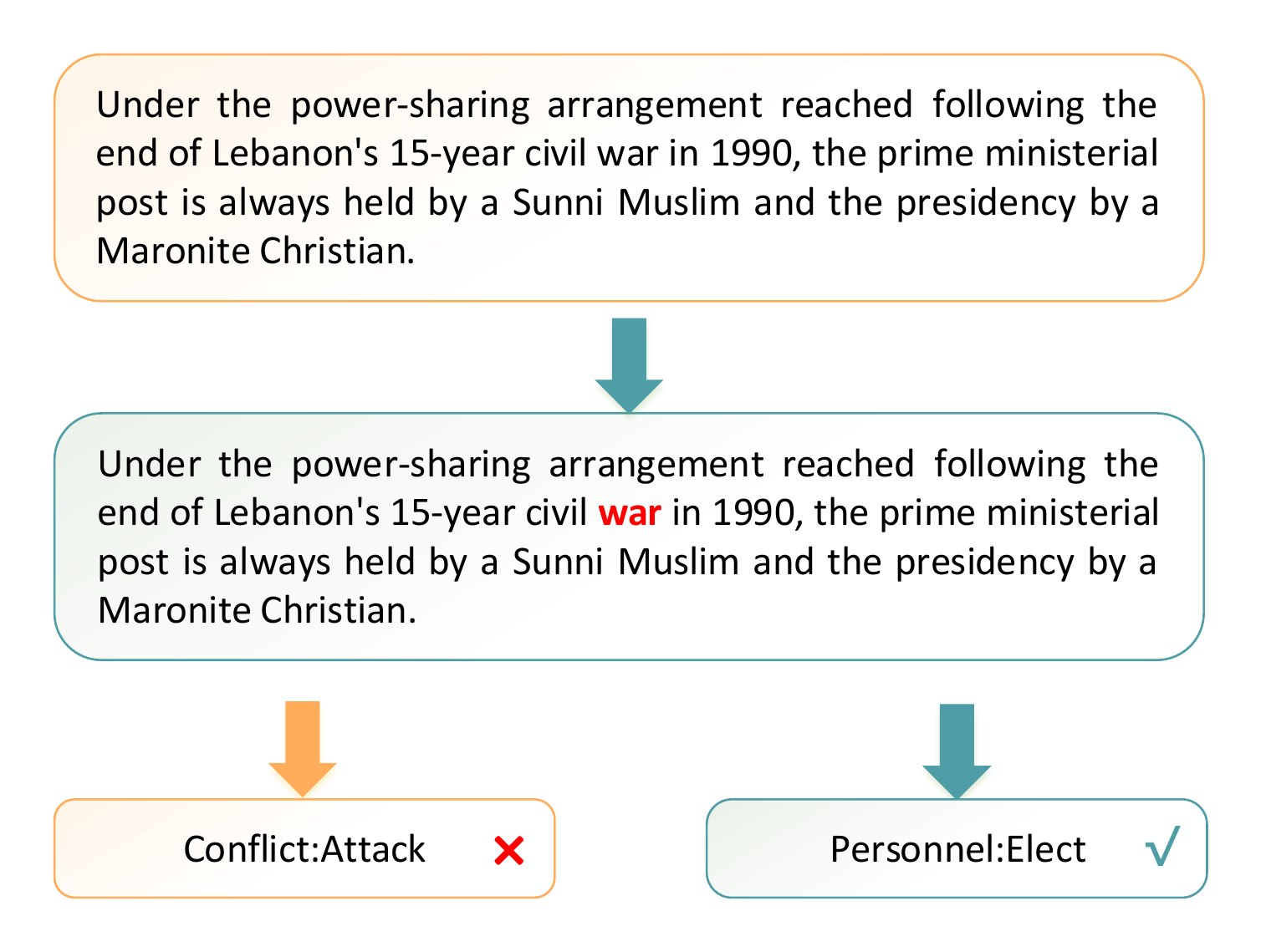}
\caption{An instance of event mention on \emph{ACE-2005}. The red marker ``\emph{war}" represents the predicted trigger word, which is also the ground-truth trigger label. The yellow arrow in the lower left corner points to the actual event label ``\emph{Conflict:Attack}", while the green arrow in the lower right corner points to the event type ``\emph{Personnel:Elect}" predicted by MsPrompt.}
\label{Figure7}
\end{figure}

To answer RQ7 and verify the ability of our model to alleviate the context-bypassing problem caused by the trigger bias, we investigate some cases where the predicted trigger words are consistent with the annotated trigger labels.

In \emph{ACE-2005}, we find that the event mentions with the trigger word ``\emph{war}" are almost all marked as the event type ``\emph{Conflict:Attack}".
However, among the event mentions that MsPrompt recognizes the trigger word is ``\emph{war}", many other sparse event labels are predicted as well in addition to the dense event type ``\emph{Conflict:Attack}".

We pick an instance from  \emph{ACE-2005} and present the event mention in Fig.~\ref{Figure7}.
Although both the trigger word predicted by MsPrompt and the ground-truth trigger word labeled manually are ``\emph{war}", the event detection results for the event type are not the same.
We can observe that the event label manually marked is still ``\emph{Conflict:Attack}", however, our model identifies the mention with the event type ``\emph{Personnel:Elect}".
Combined with the semantic understanding, it can be found that the original mention mainly describes the latter event type, which indicates MsPrompt can perform well.

In general, during the construction of event detection datasets, it is inevitable that there will be annotation inertia in human labeling of event types, i.e., habitually classifying the trigger words as the same event type.
This exposes the context-bypassing problem, which is exacerbated by the trigger bias and can lead to wrong predictions.
In contrast, our model MsPrompt pays an attention to not only the trigger word but also the original event mention in the event classifier.
Therefore, it greatly avoids this labeling inertia and mitigates the context-bypassing problem caused by the trigger bias, which confirms the debiasing effect of our proposal.

\section{Conclusion}
\label{Conclusion}

To address the data-poor dilemma and the trigger bias in event detection, we propose a novel approach MsPrompt based on the true few-shot paradigm.
We first apply an under-sampling module to adapt to the low-resource scenarios and the true few-shot setting.
Then, for the severe context-bypassing and disabled generalization caused by the trigger bias, a multi-step prompt module combined with a knowledge-enhanced ontology is employed to alleviate the context-bypassing problem.
In addition, a prototypical module is utilized to efficiently capture event label features in the low-resource occasions to further mitigate the generalization disability.
The experimental results show that our model achieves notable performance advantages, especially in the strict low-resource scenarios, and can effectively deal with the debiasing issue for few-shot event detection.

Regarding future work, on the one hand, we will consider evaluating our model in a challenging zero-shot scenario and expand to the open domain to investigate its generalization.
On the other hand, we plan to study the sensitive multi-trigger event detection models using the \emph{MAVEN} dataset \citep{emnlp20/Wang}.

\bibliographystyle{aaai}
\bibliography{refs}

\end{document}